\pdfoutput=1
\documentclass[11pt]{article}

\usepackage{ACL2023}

\usepackage{times}
\usepackage{latexsym}

\usepackage{amsmath,amssymb}
\usepackage{subcaption}
\usepackage{graphicx}
\usepackage{float}
\usepackage{stfloats}
\usepackage{fdsymbol}
\usepackage{pifont}
\usepackage{xcolor}
\usepackage{cleveref}
\usepackage{adjustbox}
\usepackage{diagbox}
\usepackage{booktabs}
\usepackage{array}
\usepackage{booktabs}
\usepackage{multirow}
\usepackage{mdframed}
\usepackage{enumitem}
\usepackage{tabularx}
\usepackage{algorithm}
\usepackage{algorithmicx}  
\usepackage{algpseudocode}  
\usepackage{diagbox}
\crefname{section}{§}{§§}
\Crefname{section}{§}{§§}
\usepackage{pifont}
\usepackage{tikz}
\usepackage{xspace}
\usepackage{expl3}

\usepackage{todonotes}
\definecolor{TodoColor}{rgb}{1,0.7,0.6}
\definecolor{TodoColor2}{HTML}{AACCAA}

\makeatletter\def\Hy@Warning#1{}\makeatother
\let\svthefootnote\thefootnote
\newcommand\blankfootnote[1]{%
  \let\thefootnote\relax\footnotetext{#1}%
  \let\thefootnote\svthefootnote%
}

\usepackage{setspace}

\usepackage[T1]{fontenc}
\usepackage[utf8]{inputenc}

\usepackage{microtype}

\usepackage{inconsolata}

\newcommand{\zoneicon}[1]{%
  \begin{tikzpicture}
    \fill[#1] (0,0) circle (0.13cm);
  \end{tikzpicture}%
}

\definecolor{z_red}{rgb}{0.8, 0.25, 0.25}
\definecolor{z_green}{rgb}{0.18, 0.55, 0.34}
\definecolor{z_orange}{rgb}{0.93, 0.57, 0.13}
\definecolor{z_blue}{rgb}{0.16, 0.32, 0.75}

\newcommand{\zonex}{\xspace \zoneicon{z_green}{$\mathcal{Z}_{\text{\ding{51}}}$}\xspace}
\newcommand{\zoney}{\xspace \zoneicon{z_blue}{$\mathcal{Z}_{\text{\ding{55}} \rightarrow \text{\ding{51}}}$}\xspace}
\newcommand{\zonez}{\xspace \zoneicon{z_red}{$\mathcal{Z}_{\text{\ding{55}} \rightarrow \text{\ding{55}}}$}\xspace}
\newcommand{\zonexx}{\xspace \zoneicon{z_orange}{$\mathcal{Z}_{\text{\ding{51}} \rightarrow \text{\ding{55}}}$}\xspace}

\ExplSyntaxOn
\NewDocumentCommand{\cb}{m}
{
  \fp_set:Nn \l_tmpa_fp { 60 } 
  \adjustbox{margin=1.5pt, bgcolor=gray!\fp_eval:n{\l_tmpa_fp*#1}}{#1}
}
\ExplSyntaxOff

\title{Investigating the Zone of Proximal Development of Language Models\\ for In-Context Learning}

\author{
Peng Cui~\qquad~Mrinmaya Sachan \\
Department of Computer Science, ETH Zürich \\
\texttt{
\href{mailto:peng.cui@inf.ethz.ch}{peng.cui@inf.ethz.ch}
}\\
}

\begin{document}
\maketitle
\begin{abstract}

In this paper, we introduce a learning analytics framework to analyze the in-context learning (ICL) behavior of large language models (LLMs) through the lens of the Zone of Proximal Development (ZPD), an %well-
established theory in educational psychology. 
ZPD delineates the space between what a learner is capable of doing unsupported and what the learner cannot do even with support.
We adapt this concept to ICL, measuring the ZPD of LLMs based on model performance on individual examples before and after ICL.
Furthermore, we propose an item response theory (IRT) model to predict the distribution of zones for LLMs.
Our findings reveal a series of intricate and multifaceted behaviors of ICL, providing new insights into understanding and leveraging this technique. 
Finally, we demonstrate how our framework can enhance LLM in both inference and fine-tuning scenarios:
(1) By predicting a model’s zone of proximal development, we selectively apply ICL to queries that are most likely to benefit from demonstrations, achieving a better balance between inference cost and performance; 
(2) We propose a human-like curriculum for fine-tuning, which prioritizes examples within the model’s ZPD. 
The curriculum results in improved performance, and we explain its effectiveness through an analysis of the training dynamics of LLMs.\footnote{Code is available at \href{https://github.com/nlpcui/llm_zpd}{https://github.com/nlpcui/llm-zpd}} 

\end{abstract}

\section{Introduction}
Human learning is a dynamic and progressive process where learners integrate new information into their knowledge base through interactions with the environment \cite{piaget1977development}.
Research in education and learning sciences has extensively explored what makes learning most effective and efficient.
Among them, the Zone of Proximal Development (ZPD) emphasizes the alignment between the learner's capability and the problem's difficulty \cite{vygotsky1978mind}.
Specifically, ZPD refers to the range of problems that a learner can solve with appropriate scaffolding but cannot tackle independently.
This concept is essential in education as
it identifies knowledge that is valuable for learning, feasible to acquire, and not yet mastered. 
Therefore, learning within ZPD is believed to foster more effective cognitive development \cite{chaiklin2003zone, tharp1991rousing}.

In this paper, we propose a learning analytics framework to study the \emph{learning behavior} of language models through the lens of ZPD. 
In particular, we focus on in-context learning (ICL), an emerging ability of LLMs that allows them to learn from a few demonstrations \cite{brown2020language,wei2022emergent}.
Previous studies have primarily focused on strategies for demonstration optimization \citep{liu2021makes,qin2023context,rubin2021learning,ye2023compositional}. 
However, even with high-quality demonstrations, the performance of ICL still varies significantly across tasks and data \cite{srivastava-etal-2024-nice}.
This variability calls for a more comprehensive examination of the \emph{inherent} in-context learnability of LLMs on individual queries. 

\begin{figure}
    \centering
    \includegraphics[width=\columnwidth]{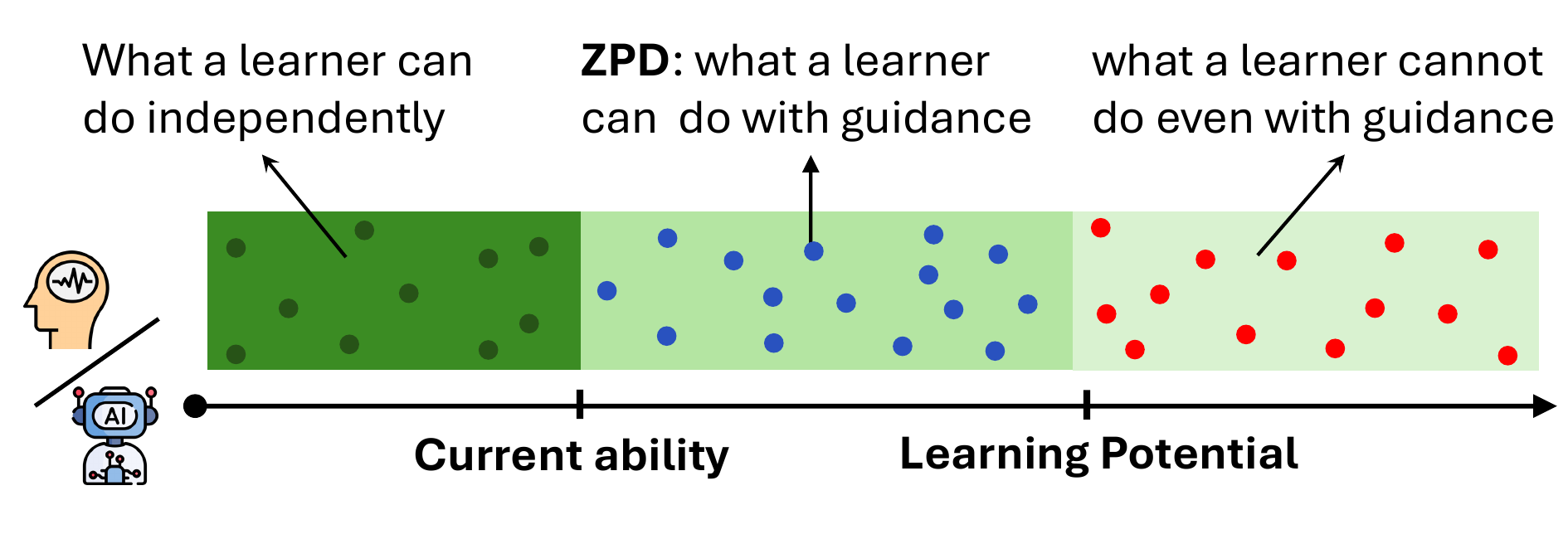}
    \caption{We conceptualize an LLM's Zone of Proximal Development (ZPD) for ICL as the set of queries on which the model's performance can be improved with demonstrations. We introduce a framework to measure and predict this zone and explore its applications.}
    \label{fig:illustration}
\end{figure}

We first formalize the concept of ZPD in ICL. 
Drawing on the parallel between ICL and human learning from worked examples, we view LLMs as learners and in-context demonstrations as a form of scaffolding. 
Then, based on the model's prior knowledge and its response to ICL, a query set can be divided into three \textbf{zones} ($\mathcal{Z}$): 
(1) The first zone, denoted as \zonex, consists of queries that can be solved by the model via direct prompting, representing the model's prior knowledge;
(2) The second zone, denoted as \zoney, includes queries that can be solved by the model only with ICL, representing the model's ZPD; and
(3) The third zone, denoted as \zonez, contains queries that the model cannot solve even with ICL, representing the knowledge beyond the model's reach. 
Figure \ref{fig:illustration} illustrates this conceptualization. 
This categorization provides a granular look at the model's capability, limitations, and interaction with specific interventions. 

We begin by measuring the task-specific zones of various models (\cref{sec: measure}). 
Since the ICL performance is sensitive to the choice of demonstrations and the ground-truth demonstrations are not available, it is non-trivial to determine whether a problem can potentially benefit from ICL.
To address this, we employ a greedy algorithm to construct \emph{Oracle} demonstrations for each query and use them to approximate the zone distribution empirically. 
Then, we propose to predict the zones of unseen queries using the item Response theory (IRT; \citet{santor1998progress}), which jointly captures the latent traits of the model and query (e.g., ability, difficulty). 
In particular, we introduce a variant of IRT that further takes into account the model's in-context learnability to capture the performance changes with or without ICL (\cref{sec: pred}).
We find that the ICL behavior of LLMs is generally predictable even without demonstration information, although the degree of predictability varies across different datasets and tasks.

Finally, we showcase how our framework enhances LLMs in both inference and fine-tuning scenarios (\cref{Sec: zone_application}). 
For inference, we propose a selective ICL strategy, which first predicts the zone of input queries and then applies ICL only to queries that are most likely to benefit from ICL (i.e., within the model’s ZPD \zoney). 
Experimental results show this approach achieves competitive or even better performance with reduced inference cost.
For fine-tuning, we propose a ZPD-based curriculum that prioritizes challenging yet learnable training examples.
We find such a curriculum improves fine-tuning outcomes.
Upon further analysis of training dynamics, we find LLMs exhibit  \emph{consistent learnability} under both ICL and fine-tuning settings.
This consistency explains the effectiveness of our ZPD-based curriculum and suggests potential connections between these two learning paradigms.

In summary, our contributions are threefold:
\begin{itemize}
 \item We conceptualize the ZPD framework for LLMs, which provides a new perspective on analyzing their ICL behavior.
 \item We introduce a novel IRT variant that captures LLMs’ in-context learnability and predicts their performance with or without ICL.
 \item We showcase two applications of our framework: a selective ICL strategy and a ZPD-based curriculum, demonstrating its potential to enhance both LLM training and inference.
\end{itemize}

\section{Related Work}
\noindent \textbf{In-Context Learning} \cite{brown2020language} has become a popular paradigm for enhancing the capabilities of LLMs across a wide range of tasks. 
Previous work has extensively focused on optimizing demonstrations, particularly through the selection \cite{liu-etal-2022-makes,rubin-etal-2022-learning,li2023unified} and ranking \cite{pmlr-v139-zhao21c,lu-etal-2022-fantastically} of in-context examples. 
In this paper, we shift the focus from demonstration optimization to the LLM and the target query themselves, highlighting the inherent in-context learnability of LLMs on individual queries. 
Our study complements these works, contributing to a holistic understanding of what makes ICL (un)successful.
Another line of research explores how the ICL capability emerges and functions, with various hypotheses proposed, such as task recognition \cite{xie2022incontext, 10.5555/3666122.3666809}, composition \cite{li-etal-2024-language}, meta-gradient learning \cite{garg2022can,akyurek2023what}. 
This paper also aims to understand ICL but from an empirical perspective by collecting, analyzing, and predicting ICL behaviors.

\vspace{1.5mm}
\noindent \textbf{Adoption of IRT in NLP.}
IRT is a set of statistical models used in educational assessments to measure the latent abilities of individuals through standardized testing \cite{lord2008statistical, santor1998progress}. 
In recent years, it has become increasingly popular in NLP. 
\citet{byrd2022predicting} uses IRT to estimate question difficulty and model
skills. 
\citet{gor2024great} proposes a content-aware and identifiable IRT to analyze human-AI complementarity. 
\citet{polo2024tinybenchmarks} argues for the adoption of IRT to build benchmarks for efficient evaluation.
In this work, we use IRT to predict LLM in-context learnability on individual queries (conceptualized as ZPD) by capturing the behavior of LLMs before and after (in-context) learning.

\vspace{1.5mm}
\noindent \textbf{Curriculum Learning} \cite{bengio2009curriculum} is the approach that organizes the training examples such that the model converges faster and better, which has been successfully applied in various NLP tasks \cite{tay2019simple,platanios2019competence,sachan2016easy}. 
Typically, curriculum learning algorithms organize training examples in increasing order of difficulty. 
Conversely, there is another line of research that works in the opposite way to start with hard examples, namely Hard Example Mining \cite{shrivastava2016training,jin2018unsupervised}. 
In this paper, we propose a ZPD-based curriculum that strikes a middle point between the two techniques: prioritizing training examples that are challenging and yet learnable (i.e., within the model's ZPD).
Similar strategies have been proven effective in various scenarios \cite{mindermann2022prioritized}. 
However, this paper proposes a new framework for discovering such desired examples, which can be incorporated into existing approaches.

\section{Measuring ZPD of LLMs}\label{sec: measure}
\subsection{Preliminaries} \label{sec: preliminary}
Let $\mathcal{D}=\{(x_1, y_1), ..., (x_n, y_n)\}$ be a dataset where $x_i$ is a query and $y_i$ is the ground-truth answer. 
We define the ZPD (\zoney) of a model $\mathcal{M}$ on $D$ as a subset of examples on which the model's performance can be improved through a learning trial. 
In this study, we focus on the ICL setting and measure learning outcomes by comparing the model's performance with and without ICL. 
Specifically, let $c = \{ (x_1, y_1)...(x_{k}, y_{k}) | x_{j} \in \mathcal{D}\}$ be a set of demonstrations for $x$ $(x \notin c)$, we define \zoney as:
\begin{gather}
    \text{\zoney} \triangleq \{x | \mathcal{F}(y^\varnothing)<\tau, \mathcal{F}(y^{c}) > \tau \},
\end{gather}
where $\mathcal{F}$ is a scoring function and $\tau$ is a threshold deciding whether the predicted answer is acceptable. 
$y^{\varnothing}$ and $y^{c}$ represent the model's output with direct prompting and with in-context demonstrations, respectively:
\begin{gather}
    y^{\varnothing} = \mathcal{M}(\mathcal{T}(x)), y^{c} = \mathcal{M}(\mathcal{T}(c_1) \oplus ... \oplus \mathcal{T}(x)).
\end{gather}
where $\mathcal{T}$ is a template function and $\oplus$ denotes string concatenation. 
Due to the potential interference between instruction and demonstration \cite{srivastava-etal-2024-nice}, we adopted a simple prompt template with minimal instruction to focus on the effect of demonstration (See Appendix Table \ref{tab:app_prompt}).
 
Similarly, we can define the other two subsets as follows:
\begin{gather}
    \text{\zonex} \triangleq \{x | \mathcal{F}(y^\varnothing)>\tau \}, \\
    \text{\zonez} \triangleq \{x | \mathcal{F}(y^\varnothing)<\tau, \mathcal{F}(y^{c})<\tau \},
\end{gather}
representing queries that can be solved by $\mathcal{M}$ with direct prompting,
and queries that cannot be solved even with ICL. 

This formalization is flexible and can be applied to other settings.
For example, future work could replace ICL with other prompting strategies or analyze fine-tuning behaviors by examining the performance across different epochs. 

\subsection{Approximating \zoney and \zonez}
While \zonex is deterministic from the model's base performance $\{y^{\varnothing}_1, y^{\varnothing}_2, ...\}$, \zoney and \zonez depend on the choice of demonstrations $c$.
In this paper, we aim to investigate the ideal ICL behavior of LLMs with \emph{optimal} demonstrations.
This is because our goal is to understand the model’s \emph{inherent} in-context learnability on individual queries rather than the behavior of a specific ICL strategy. 
Since optimal demonstrations for each query are unavailable, precise measurements of \zoney and \zonez are infeasible. 
To address this, we first create \emph{Oracle} demonstrations---the best demonstrations achievable in a practical setting (with a limited demonstration pool and restricted computation resources).
Then, we use them to approximate \zoney and \zonez.

In concrete, we adopt a retrieve and rank method to construct Oracle demonstrations.
Firstly, we retrieve a candidate set $\mathcal{C}$ for each query. 
The common belief is that demonstrations that are similar to the query are most likely to enhance performance \cite{liu-etal-2022-makes}. 
Following previous work \cite{rubin-etal-2022-learning}, we employ BM25 \cite{robertson2009probabilistic}, a sparse retriever based on surface features, and SBERT \cite{reimers2019sentence}, which is based on dense sentence encoding. 
For the two retrievers, we calculate similarities based on both the $(x, y)$ pair and the ground-truth answer $y$ only, resulting in 2 $\times$ 2 $\times$ $K$ candidates.  
However, similarity may not be the only criterion for demonstration selection. 
To further enrich the candidate set and recall effective but dissimilar demonstrations, we randomly sample $K$ candidates from the bottom 50 percentile of the retrieving results, doubling the candidate size. 

Next, we select Oracle demonstrations $c$ using a greedy scoring approach:
\begin{gather}
    c_i = \mathop{{\rm argmax}}\limits_{\mathcal{C} \setminus \{c_1,..,c_{i-1}\} } {\rm Prob}_{\mathcal{M}} (y|c_1 \oplus ... c_{i} \oplus x),
\end{gather}
where $c_i$ is the $i^{th}$ selected demonstration and ${\rm Prob_{\mathcal{M}}(\cdot)}$ is the probability from the model $\mathcal{M}$. 
In other words, we greedily choose demonstrations that can maximize the likelihood of the ground-truth answer.
With these demonstrations, the resulting \zoney is a subset of the actual ZPD while \zonez is a superset of the actual one.
In the rest of the paper, we use \zoney and \zonez to denote for the approximated zones unless otherwise specified. 

\section{Zone Prediction} \label{sec: pred}
\begin{figure}[t]
    \centering
    \includegraphics[width=\columnwidth]{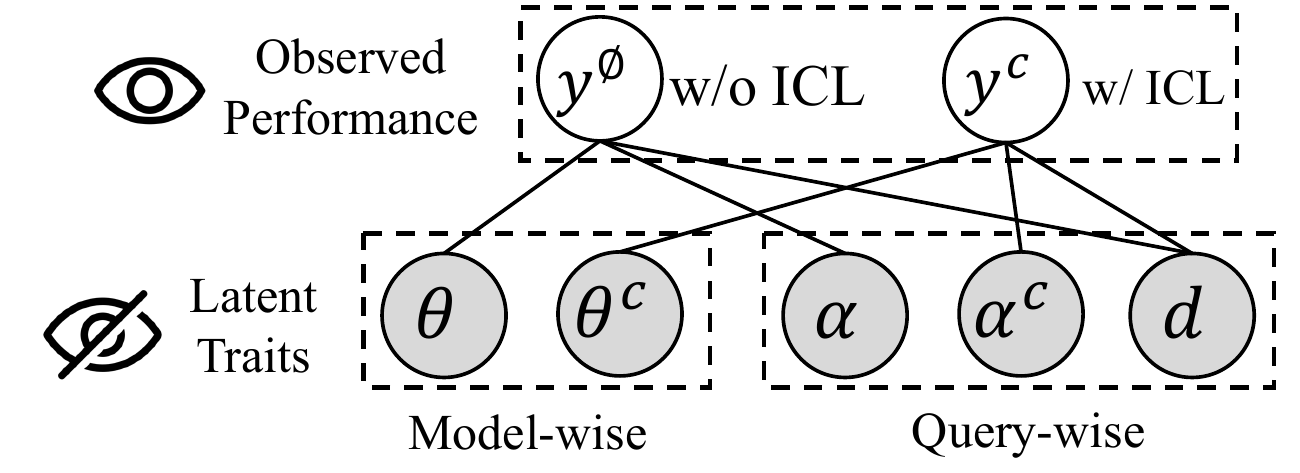}
    \caption{
    We assume that a model's performance on a given query, $y^{c}$ (with ICL) or $y^{\varnothing}$ (without ICL), is determined by latent traits (shadowed nodes, bottom) of both the model and the query, including the model's skill $\boldsymbol{\theta}$, ICL skill $\boldsymbol{\theta^{c}}$, the query's discrimination $\boldsymbol{\alpha}$, ICL discrimination $\boldsymbol{\alpha^c}$, and overall difficulty $d$. 
    }
    \label{fig:model}
\end{figure}

In this section, we attempt to build a model to predict an LLM’s zone distribution on unseen queries. 
Essentially, the goal is to predict the model's performance, i.e., whether it can solve a query directly (\zonex) or with ICL (\zoney), or not at all (\zonez). 
We propose a novel variant of item response theory (\textsc{Irt}) to capture the latent traits of the LLM and the queries.
A graphic view of our model is shown in Figure \ref{fig:model}.

\subsection{Background of \textsc{Irt}}
\textsc{Irt} is a statistical model that predicts the probability of individual respondents correctly answering a set of queries (or items).
In this work, we take a collection of LLMs $\{\mathcal{M}_1, \mathcal{M}_2, ..., \mathcal{M}_m\}$ as respondents. 
The basic 1 Parameter Logistic (1PL) \textsc{Irt} is defined as:
\begin{gather}
    P(r_{i,j}=1|\mathcal{M}_{i}, x_{j}) = \sigma(\theta_{i}-d_{j}), \label{eq:irt_1PL}
    \end{gather} 
where $r_{i,j}$ is the binary correctness label of $\mathcal{M}$'s prediction on $x_i$. 
$\sigma$ is the ${\rm sigmoid}$ function. 
$\theta_i$ and $d_j$ are latent variables (scalars) to be estimated, representing the ability of the $i$th model $\mathcal{M}_{i}$ and the difficulty of the $j$th query $x_j$. 
Simply put, \textsc{Irt} predicts the correctness label based on the gap between model ability and query difficulty.

The 1PL \textsc{Irt} assumes the monotonic relationship between item difficulty and respondent ability.
To relax this, we employ the multi-dimensional IRT (\textsc{MIrt}, \citet{reckase200618}), which is defined as:
\begin{gather}
    P(r_{i,j}=1|\mathcal{M}_{i}, x_{j}) = \sigma(\boldsymbol{\theta}_{i}^{\mathrm{T}}\boldsymbol{\alpha}_{j}-d_{j}), \label{eq:MIRT-DP}
\end{gather}
where the model's ability is represented as a \emph{skill vector} $\boldsymbol{\theta}_j \in \mathbb{R}^{\rm H}$.
Correspondingly, an item-wise \emph{discrimination vector} $\boldsymbol{\alpha}_i \in \mathbb{R}^{\rm H}$ is introduced to represent its latent traits.
A closer alignment between $\boldsymbol{\theta}_{i}$ and $\boldsymbol{\alpha}_j$ indicates a higher likelihood of a correct response.
%\mrinmaya{I think the above part about MIRT can be written better. We}

The training objective of \textsc{Irt} is defined as:
\begin{gather}
    \mathcal{L}_{\rm IRT}=\sum_{i=1}^{\rm M}\sum_{j=1}^{\rm N} {\rm CE}(P(r_{i,j}), y_{j}),
\end{gather}
where ${\rm CE(\cdot)}$ stands for the cross-entropy loss between predicted probability and the groud-truth label.

\subsection{Content-Aware \textsc{Mirt}}
A limitation of \textsc{MIrt} is that it relies on the response data to infer item traits $\boldsymbol{\alpha}_i$. 
Therefore, it cannot generalize to unseen queries during inference.
To overcome this limitation, we use a lightweight neural network to parameterize item traits based on their text features. 
Specifically, for a given query $x_{j}$, we first use an embedding model to obtain its representation $\boldsymbol{e}_j$.
Then, we compute its traits by:
\begin{gather}
    d_{j} = f({\rm {\bf W}_{d}} \boldsymbol{e}_{j} + {\rm {\bf b}_{d}}); 
    \boldsymbol{\alpha}_j = f({\rm {\bf W}_{\alpha}} \boldsymbol{e}_{j} + {\rm {\bf b}_{\alpha}})
\end{gather}
where ${\rm {\bf W}_{d}}, {\rm {\bf W}_{\alpha}}, {\rm {\bf b}_{d}, {\rm {\bf b}_{\alpha}}}$ are learnable weights, trained together with the IRT model, and $f$ is the ${\rm Relu}$ function.

\subsection{Adapting \textsc{Mirt} to Learning Dynamics}
While the above model can predict the model's performance on an unseen query, it cannot predict one query's correctness label under two settings and thus cannot predict three zones simultaneously.
We propose a variant that incorporates the dynamics of ICL. 
Concretely, we introduce an additional \emph{ICL skill vector} $\boldsymbol{\theta}^{c}$ for the model and similarly an \emph{ICL discrimination vector} $\boldsymbol{\alpha}^{c}$ for the item:
\begin{gather}
         P(r_{i,j}=1|\mathcal{M}_{i}, x_{j}) = \sigma(\boldsymbol{\theta}_{i}^{\mathrm{T}}\boldsymbol{\alpha}_j-d_{j}+\boldsymbol{\theta}^{c\mathrm{T}}_{i }\boldsymbol{\alpha}_{j}^{c}), \label{eq:MIRT-ICL}
\end{gather}
where the alignment between $\boldsymbol{\theta}^{c}_{i}$ and $\boldsymbol{\alpha}_{j}^{c}$ represents the in-context \emph{learnability} of $\mathcal{M}_{i}$ with respect to $x_j$. 
Similar to $d$ and $\boldsymbol{\alpha}$, $\boldsymbol{\alpha}^{c}_{j}$ is computed based on the embedding of x:
\begin{gather}
    \alpha^{c}_{j} = f({\rm {\bf W}^{c}_{\alpha}} e_{j} + {\rm {\bf b}^{c}_{\alpha}}).
\end{gather}
Combining Eq. \ref{eq:MIRT-DP}, and \ref{eq:MIRT-ICL}, we have:
\begin{gather}
         P(r_{i,j}^{\{ \varnothing, c \}}=1) = \sigma(\boldsymbol{\theta_{i}} \boldsymbol{\alpha_j} - d_{j} + g^{\{ \varnothing, c \} } \boldsymbol{\theta}^{c}_{i }\boldsymbol{\alpha}_{j}^{c}), \label{eq:combined}
\end{gather}
where $r^{\varnothing}$ and $r^{c}$ are the correctness labels under direct prompting and ICL.
$\{g^\varnothing=0, g^c=1\}$ is a gating parameter in align with $r$ to ensure that $\theta^{c}_{i }\alpha_{j}^{c}$ are only enabled in the ICL setting. 
In doing this, the latent factors are learned such that:
\begin{gather}
    \left\{
        \begin{array}{cc}
            \boldsymbol{\theta}^{\mathrm{T}} \boldsymbol{\alpha} > d, \boldsymbol{\theta}^{\mathrm{T}} \boldsymbol{\alpha} + \boldsymbol{\theta}^{c\mathrm{T}} \boldsymbol{\alpha}^{c} >d, \ \text{if} \ \ r^{\varnothing}=1, &\\
             \boldsymbol{\theta}^{\mathrm{T}} \boldsymbol{\alpha} <d, \boldsymbol{\theta}^{\mathrm{T}} \boldsymbol{\alpha} + \boldsymbol{\theta}^{c\mathrm{T}} \boldsymbol{\alpha}^{c} >d, \ \text{if} \ \ r^{\varnothing}=0, r^{c}=1,  &\\
             \boldsymbol{\theta}^\mathrm{T} \boldsymbol{\alpha} <d, \boldsymbol{\theta}^\mathrm{T} \boldsymbol{\alpha} + \boldsymbol{\theta}^{c} \mathrm{T} \boldsymbol{\alpha}^{c} <d, \ \text{if} \ \ r^{\varnothing}=0, r^{c}=0. &
        \end{array}
    \right.
\end{gather}
The above three situations correspond to \zonex, \zoney, and \zonez, respectively. 
We refer to the proposed model as \textsc{Mirt}$_{\textsc{Icl}}$. 

From a multi-task learning perspective, our model can be seen as jointly training two IRT models, each with its own ability ($\theta, \theta^{c}$) and discrimination ($\alpha, \alpha^{c}$) parameters, while sharing the overall item difficulty ($d$). 
This allows the model to better capture the relationships between LM behaviors across the two settings.

\section{Experiments}
We experiment with 8 LLaMA models \cite{touvron2023llama,dubey2024llama} of various sizes, including \texttt{LLaMA-2-7B}, \texttt{LLaMA-2-7B-chat},
\texttt{LLaMA-2-13B}, \texttt{LLaMA-2-13B-chat}, \texttt{LLaMA-3-8B}, \texttt{LLaMA-3-8B-Instruct},  \texttt{LLaMA-3-70B}, and \texttt{LLaMA-3-70B-Instruct}. 
In particular, we consider both instruction-tuned (IT) (\texttt{-chat/Instruct} models) or non-IT versions to examine the influence of instruction tuning on the model’s ZPD.
In this study, we focus on the \emph{mathematical reasoning} and \emph{text understanding} abilities of LLMs, using the MathQA dataset \textbf{GSM8K} \cite{cobbe2021gsm8k} and the Stance detection (Favor, Neutral, Against) dataset \textbf{EZStance}  \cite{zhao2023ez} for stance detection. 
Detailed experiment setup can be found in Appendix \ref{sec:app_setup}.

We first present and analyze the zone distribution of various LLaMA models (\cref{Sec:zone_dist}).
Then, we evaluate the performance of IRT models on zone prediction (\cref{Sec: zone_pred}). 
Finally, we demonstrate two applications of our framework (\cref{Sec: zone_application}). 

\subsection{Zone Distribution Analysis} \label{Sec:zone_dist}
\begin{figure}
    \centering
    \includegraphics[width=\columnwidth]{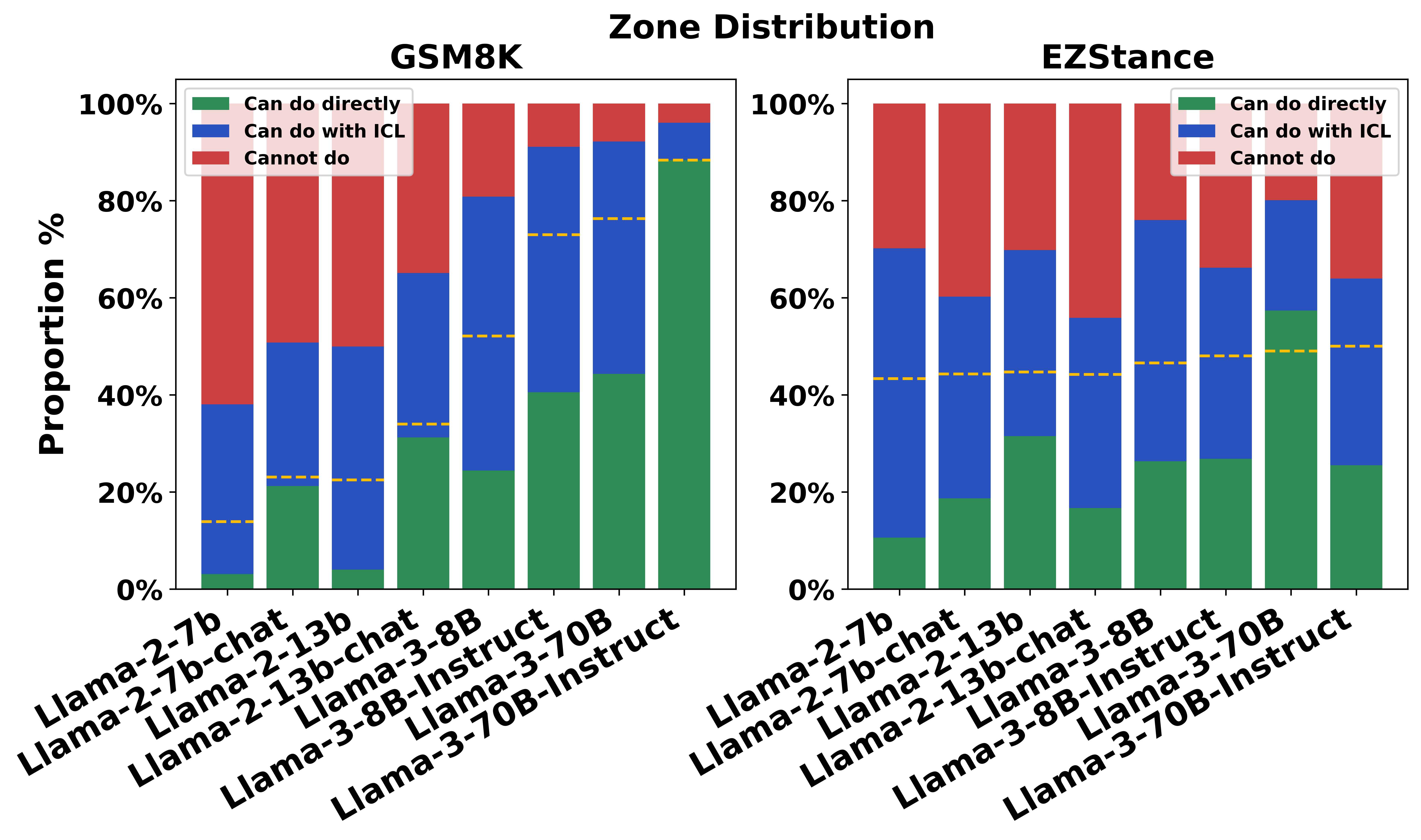}
    \caption{Zone distribution of various LLMs on the two datasets. Yellow lines represent the accuracy of \textsc{Kate}.}
    \label{fig:zone_dist}
\end{figure}
We measure the three zones of LLaMA models on the test set of GSM8K and the validation set of EZStance. 
Our observations are as follows.

\vspace{1.5mm}
\noindent \textbullet\ \emph{\textbf{The potential of ICL remains largely untapped}}. 
In Figure \ref{fig:zone_dist}, we present the zone distributions of various models. 
Ideally, the accuracy of ICL should be the combined proportion of \zonex and \zoney, which highlights the great potential of ICL. 
For instance, on the GSM8K dataset, the 8B-Instruct model, with the help of Oracle demonstrations, can achieve competitive performance compared to the two 70B models.
Note that, however, this is only a lower bound of ideal ICL performance, as the Oracle demonstrations are still sub-optimal.
Nevertheless, the current method still falls short of fully utilizing even this lower bound. 
For reference, we highlight the accuracy (yellow line) of \textsc{Kate} \cite{liu2021makes}, a similarity-based demonstration selection strategy (with \texttt{paraphrase-mpnet-base-v2}).
On average, it lags by around 20\% on the two datasets.

\begin{figure}[t]
    \centering
    \includegraphics[width=\columnwidth]{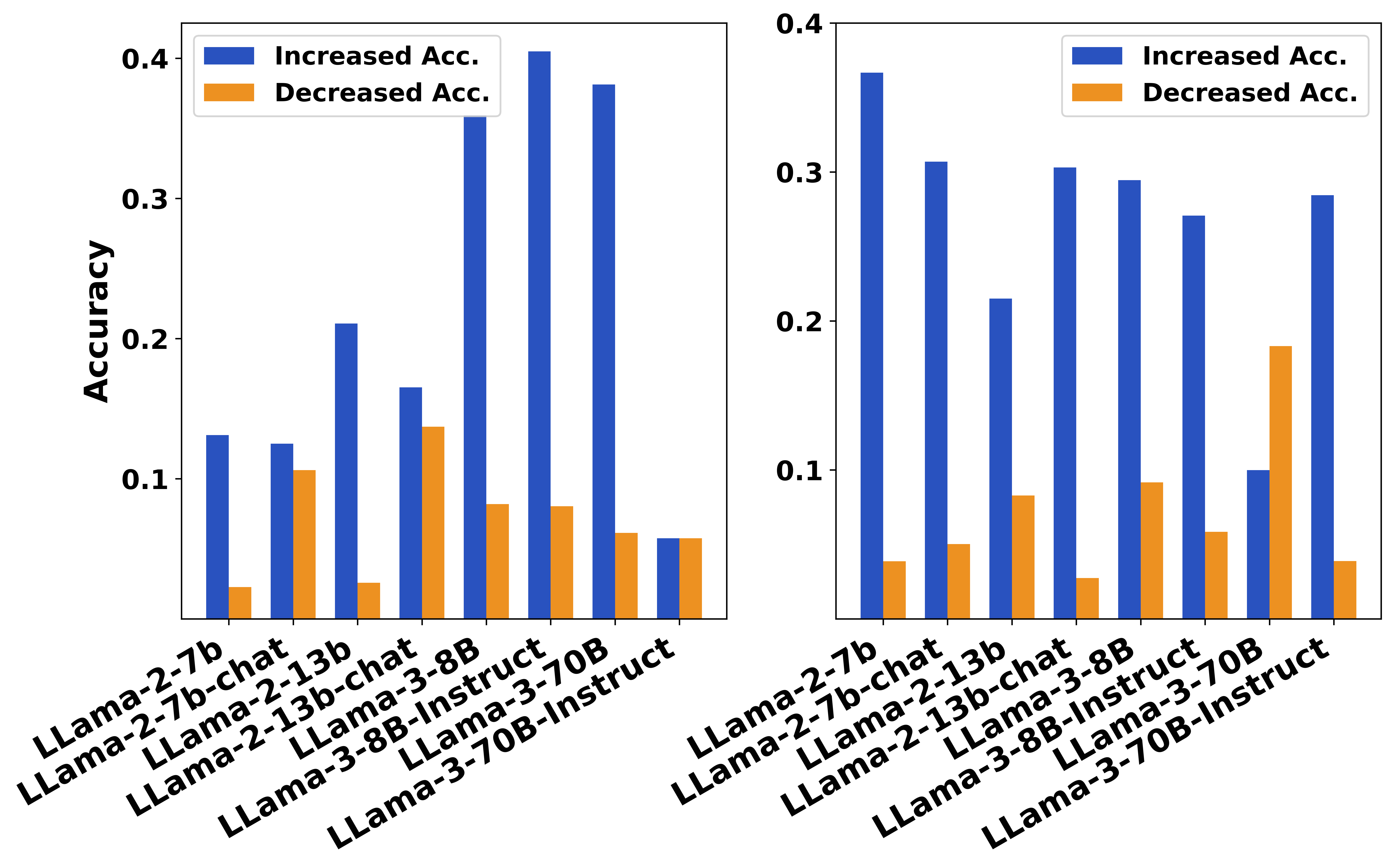}
    \caption{Increased and decreased accuracy by \textsc{Kate} on GSM8K (left) and EZStance (right).}
    \label{fig:neg_icl}
\end{figure}

\vspace{1.5mm}
\noindent \textbullet\ \emph{\textbf{In-context demonstrations can be harmful.}}
In Section \cref{sec: measure}, we divide a query set into three zones according to the model's performance difference with and without ICL. 
However, sometimes, ICL can also degrade the performance.
We denote the collection of such queries as \zonexx. 
We do not frame \zonexx into our formalization (\cref{sec: preliminary}) but merge it into \zonex because we focus on the ideal ICL behavior given Oracle demonstrations.
However, this negative effect of ICL is non-negligible in a practical setting.

With \textsc{Kate} as a case study, we compare its increased accuracy (i.e., the proportion of recalled ZPD (\zoney) examples) and decreased accuracy (i.e., the proportion of \zonexx examples) in Figure \ref{fig:neg_icl}. 
The sum of the two is the overall performance of $\textsc{Kate}$. 
We can see \zonexx can reduce up to 14\% and 18\% accuracy on GSM8K and EZStance.
Besides, this negative effect is also model-dependent. 
For example, \texttt{LLaMA-2-7B-chat} and \texttt{LLaMA-2-13B-chat} are particularly vulnerable to harmful demonstrations, and this negative effect even overwhelms the benefit for \texttt{LLaMA-3-70B}.
% More investigations are needed to understand this behavior, which we leave for future work. 
This granular look at the ICL performance provides a new perspective to improve ICL strategy: recalling examples in \zonex while minimizing \zonexx. 
Previous work mainly focused on the first direction and we will showcase how our IRT model can enhance ICL through the second way in \cref{App1}.

\begin{table}[h]
\small
\centering
\begin{tabular}{@{}lllllll@{}}
\toprule
\multirow{2}{*}{\textbf{Zones}} & \multicolumn{3}{c}{\textbf{GSM8K}}         & \multicolumn{3}{c}{\textbf{EZStance}}      \\ \cmidrule(l){2-7} 
                                & \textbf{Max} & \textbf{Min} & \textbf{Avg} & \textbf{Max} & \textbf{Min} & \textbf{Avg} \\ \midrule
\zonex                              & \cb{0.89}         & \cb{0.74}         & \cb{0.84}         & \cb{0.91}             & \cb{0.46}              & \cb{0.70}             \\
\zoney                              & \cb{0.74}         & \cb{0.21}         & \cb{0.58}         &  \cb{0.78}            & \cb{0.34}             & \cb{0.58}             \\
\zonez                              & \cb{0.58}         & \cb{0.20}         & \cb{0.42}         &   \cb{0.87}           &  \cb{0.32}            &  \cb{0.53}            \\ \bottomrule
\end{tabular}
\caption{Pairwise overlap coefficients among zones of different LLMs.} 
\label{tab: overlap}
\end{table}

\vspace{1.5mm}
\noindent \textbullet\ \emph{\textbf{ZPD (\zoney) of LLMs differ significantly}}. 
We measure the overlap between zones of different models by calculating their averaged pairwise \emph{Overlap Coefficient}, defined as:
\begin{gather}
    \textsc{Overlap} (A, B) = \frac{|A \cap B|}{{\rm min}(|A|, |B|)},
\end{gather}
where $A$ and $B$ are the zones to compare. 
The results are shown in Table \ref{tab: overlap}, where we can see examples in \zonex are largely shared across various models, while examples in \zoney and \zonez do not highly overlap, indicating each LLM has its own ZPD.
This suggests that ICL strategies should take into account both the data aspect (e.g., similarity) and the model, corroborating the conclusion of \citet{peng-etal-2024-revisiting}.

\subsection{Zone Prediction Evaluation}\label{Sec: zone_pred}

We compare our proposed IRT model \textsc{Mirt}$_{\rm ICL}$ (Eq. \ref{eq:combined}) with the following baselines:
i) \underline{1PL model} (\textsc{Irt}$_{\rm 1PL}$, Eq. \ref{eq:irt_1PL}), 
ii) \underline{2PL model}, which is similar to Eq. \ref{eq:MIRT-DP} but with $\theta$ and $\alpha$ as scalars, 
and iii) \underline{Multi-Dimensional IRT} \textsc{Mirt} (Eq. \ref{eq:MIRT-DP}).
We evaluate their ability to predict LLM performance under both direct prompting (DP) and ICL, using AUC as the primary metric. See Appendix \ref{sec:app_irt_implementation} for the implementation details. 
Note that aside from our \textsc{Mirt}$_{\textsc{Icl}}$, other baseline models are trained solely on DP data. 
Nevertheless, we can assess their generalization ability to the ICL setting: since AUC assesses the relative ranking of predicted probabilities, these models should also achieve good AUC if \emph{the LLM's probabilities of correctly answering individual queries are consistent across both settings}.

\begin{table}[t]
\small
\centering
\setlength\tabcolsep{4pt}
\begin{tabular}{@{}lcccccc@{}}
\toprule
\multirow{2}{*}{\textbf{Model}} & \multicolumn{3}{c}{\textbf{GSM8K}} & \multicolumn{3}{c}{\textbf{EZStance}} \\ \cmidrule(l){2-7} 
                                & DP        & ICL       & Overall    & DP         & ICL        & Overall     \\ \midrule
\textsc{Irt}$_{\rm 1PL}$                             & 0.808     & 0.769     & 0.748      & 0.736      & 0.617      & 0.644       \\
\textsc{Irt}$_{\rm 2PL}$                             & 0.788     & 0.740     & 0.728      & 0.739      & 0.631      & 0.651       \\
\textsc{Mirt}                            & \textbf{0.837}     & 0.770     & 0.743      & 0.760      & 0.608      & 0.799       \\
\textsc{Mirt}$_{\rm ICL}$                            & 0.833     & \textbf{0.821}     & \textbf{0.862}      & \textbf{0.770}      & \textbf{0.662}      & \textbf{0.799}       \\ \bottomrule
\end{tabular}
\caption{Performance (AUC) of various IRT models on the two datasets. The best results are in \textbf{bold}. Results of Accuracy can be found in Appendix Table \ref{tab:app_irt_acc}.}
\label{tab:irt_auc}
\end{table}

\vspace{1.5mm}
\noindent \textbullet\ \emph{\textbf{ICL behavior is, to varying degrees, predictable without demonstrations}}. 
We present the AUC results in Table \ref{tab:irt_auc}. 
As a demonstration-agnostic model, \textsc{Mirt}$_\textsc{\ Icl}$ achieves reasonably decent performance GSM8K but comparatively weaker results on EZStance. 
We interpret the difference through the \emph{predictability} and \emph{sensitivity} of ICL: for certain tasks and datasets, ICL performance may hinge more on the model’s inherent ICL capacity and the query's difficulty. 
While for others, it may depend more on the demonstrations or prompts, making the ICL behavior less predictable without the information of demonstrations.
Existing work has been focusing on measuring and mitigating sensitivity \cite{zhao2021calibrate}. 
We highlight a complementary perspective: measuring and leveraging (See \cref{Sec: zone_application} for applications) the predictability of ICL behavior.
\vspace{1.5mm}

\begin{table}[t]
\small
\centering
\setlength\tabcolsep{2pt}
\begin{tabular}{@{}lcccccccc@{}}
\toprule
\multirow{2}{*}{} & \multicolumn{2}{c}{\textbf{L2-7B}} & \multicolumn{2}{c}{\textbf{L2-13B}} & \multicolumn{2}{c}{\textbf{L3-8B}} & \multicolumn{2}{c}{\textbf{L3-70B}} \\
                  & base    & chat   & base    & chat    & base    & instr.   & base    & instr.    \\ \midrule
\textbf{GSM8K}    & $+$.10            & $+$.40            & $+$.13$^*$             & $+$.24             & $+$.29             & $+$.07            & $+$.31             & $+$.53             \\
\textbf{EZStance} & $-$.45            & $-$.60           & $-$.47            & $-$.25            & $-$.35            & $-$.28           & $-$.48            & $-$.36            \\ \bottomrule
\end{tabular}
\caption{Pearson Correlation between $\boldsymbol{\theta}^{\mathrm{T}}\boldsymbol{\alpha}-d$ (model's ability to solve the query with direct prompting) and $\boldsymbol{\theta}^{c\mathrm{T}} \boldsymbol{\alpha}^{c}$ (the additional gain obtained by ICL). Results with $^{*}$ indicate $p$-value$>0.05$.} \label{tab:pearson}
\vspace*{-2mm}
\end{table}

\noindent \textbullet\ \emph{\textbf{(In)consistency between difficulty and in-context learnability.}}
In Eq. \ref{eq:combined}, $\boldsymbol{\theta}\boldsymbol{\alpha} - d$ represents the model's ability to solve the query with DP (or the query's \textit{difficulty}), while $\boldsymbol{\theta}^{c} \boldsymbol{\alpha}^c$ captures the additional gain achieved through ICL, reflecting the model's \textit{in-context learnability} of the query.
To examine the relationship between the two terms, we compute their Pearson correlation. 
The results, presented in Table \ref{tab:pearson}, reveal that for the GSM8K dataset, these two terms exhibit weak or moderate positive correlations (from $+$0.07 to $+$0.53). 
Interestingly, the correlation on EZStance is stronger but negative, meaning difficult examples under direct prompting (lower $\boldsymbol{\theta}^{\mathrm{T}}\boldsymbol{\alpha} - d$) seem to benefit more from ICL (higher $\boldsymbol{\theta}^{c \mathrm{T}} \boldsymbol{\alpha}^c$) and vice versa. 
% However, due to the weaker performance on EZstance, this phenomenon should be interpreted with caution. 
This suggests that a query’s difficulty and its in-context learnability are not always aligned. We attribute this phenomenon to the differing abilities required for direct prompting versus ICL. 
The former primarily relies on the model’s prior knowledge of the query, while the latter depends on its ability to leverage contextual information. 
As a result, this inconsistency could arise in certain tasks and queries where the knowledge is missing but easy to learn in context. 
A notable example is classification with flipped or semantically unrelated labels \cite{wei2023larger}, where an LM struggles to solve the disrupted task in the regular setting but can successfully learn the new mapping through demonstrations.

\subsection{Applications} \label{Sec: zone_application}
In this section, we demonstrate how our framework can improve in-context learning through a selective ICL strategy (\cref{App1}) and a ZPD-derived curriculum for fine-tuning LLMs (\cref{App2}).

\subsubsection{Selective ICL}\label{App1}
\textbf{Approach.}
While ICL has demonstrated effectiveness across a wide range of tasks, it costs $k$ times additional input tokens ($k=$ the number of demonstrations).
Moreover, as discussed in \cref{Sec:zone_dist}, ICL sometimes results in worse performance, even with carefully retrieved demonstrations.
To address these issues, we propose Selective ICL (\textsc{SelIcl}). 
In specific, given a query $x_i$, we first predict its correct probability with direct prompting $p_{i}^{\varnothing}$ and the correct probability with ICL $p_{i}^{c}$ using Eq. \ref{eq:combined} with $g=0$ and $g=1$ respectively. 
Then, we decide the inference prompt for $x_i$ by:
\begin{gather}
    \left\{
        \begin{array}{ll}
             \mathcal{T}(\tilde{c}_1) ...  \oplus \mathcal{T}(x) \:  & \text{if} \; p^{\varnothing} < \tau_1 \; \text{and} \; p^{c} > \tau_2  \\
            \mathcal{T}(x) &\text{Otherwise.} 
        \end{array}
    \right. \label{eq:sel_icl}
\end{gather}
where $\{ \tilde{c}_{1}, ..., \tilde{c}_{n} \}$ are demonstrations retrieved by a certain strategy.
$\tau_1$ and $\tau_2$ are predefined thresholds.
A lower $p^{\varnothing}$ ($< \tau_1 $) and a higher $p^{c}$ ( $>\tau_2$) indicate the model is unable to solve this query with direct prompting but is likely to solve it with ICL. 
In other words, we apply ICL only to queries within the model's ZPD. 
By doing so, we aim to reduce unnecessary costs by avoiding ICL for either too easy ($p^{\varnothing} < \tau_1 $) or too hard ($p^{c} >\tau_2$) queries.
Furthermore, this can also potentially improve performance by mitigating the negative effect of ICL observed in Figure \ref{fig:neg_icl}.    

\vspace{1.5mm}

\begin{figure*}[th]
    \centering
    \includegraphics[width=\textwidth]{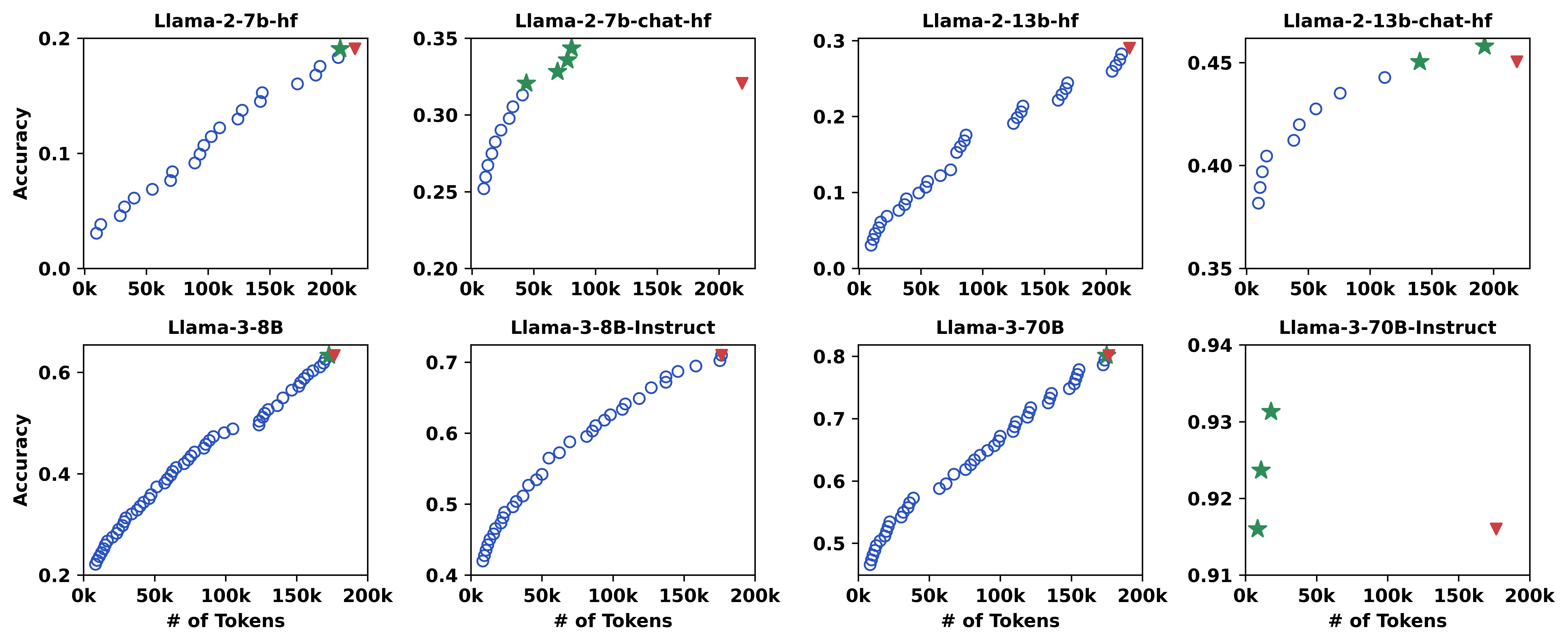}
    \caption{Accuracy and inference cost (number of input tokens) of different ICL strategies on the GSM8K dataset. \textcolor{z_red}{$\blacktriangledown$} is the performance of the baseline \textsc{FulIcl}, which applies ICL to all the queries. \textcolor{z_blue}{$\bigcirc$} and \textcolor{z_green}{$\bigstar$} are the performance of \textsc{SelIcl} under various thresholds $\tau_1$ and $\tau_2$ (not shown), where \textcolor{z_green}{$\bigstar$} highlights cases in which \textsc{SelIcl} achieves better or equal accuracy with less input tokens compared to the baseline (\textcolor{z_red}{$\blacktriangledown$}).}
    \label{fig:gsm8k_sel_icl}
\end{figure*}

\noindent \textbf{Result and Analysis.}
We compare our \textsc{SelIcl} with the vanilla ICL that applies demonstrations to all queries  (denoted as \textsc{FulIcl}). 
Specifically, we use \textsc{Kate} to retrieve demonstrations for \textsc{FulIcl}.
However, it is worth noting that \textsc{SelIcl} is orthogonal to other ICL strategies for two reasons: 
(1) It focuses on determining when to apply ICL, independent of how demonstrations are selected or organized; 
(2) The IRT model is trained to predict the model’s ICL performance given Oracle demonstrations. 
Consequently, $p^{c}$ is expected to serve as the predicted upper bound for any ICL strategy.

To select $\tau_{1}$ and $\tau_{2}$ for \textsc{SelIcl}, we perform a grid search on the IRT validation set by varying their values within the range $[0.01, 0.02, \dots, 0.99]$.
For each combination, we decide whether or not to apply ICL to each query according to Eq. \ref{eq:sel_icl} and compute the overall accuracy and number of input tokens.  
Since the prompts and model outputs are already collected when constructing the IRT dataset (Appendix \ref{sec:app_irt_implementation}), these results can be obtained without additional model inference.

Then, we plot the Pareto curve \cite{deb2011multi} of \textsc{SelIcl}, approximated with scatter points.
In multi-objective optimization, each point on the Pareto curve represents a Pareto-optimal solution that cannot be further improved in one objective without compromising the other (in our case, accuracy and number of input tokens).

Results for GSM8K are shown in Figure \ref{fig:gsm8k_sel_icl}, and results for EZstance are available in Appendix Figure \ref{fig:ezstance_sel_icl}.
Solutions that are dominated\footnote{
In the context of a Pareto curve, a solution dominates another if it is at least as good in all objectives and strictly better in at least one objective.} by others are discarded (apart from the baseline results (\textcolor{z_red}{$\blacktriangledown$}) for comparison).
As can be seen, for 6 out of 8 models, \textsc{SelIcl} with proper thresholds (\textcolor{z_green}{$\bigstar$}) can dominate \textsc{fulIcl}. 
Overall, \textsc{SelIcl} can serve as a tool to trade off accuracy and cost in resource-limited scenarios. 
\textsc{SelIcl} is paticularly successful for \texttt{LLaMA-2-7b-chat} and \texttt{LLaMA-70B-Instruct}. 
Combining with previous findings, both models have relatively narrow ZPD (Figure \ref{fig:zone_dist}) and are more susceptible to the negative effects of ICL (Figure \ref{fig:neg_icl}), suggesting that greater caution is needed when applying ICL to them.

\subsubsection{ZPD-based Curriculum}\label{App2}
It is generally believed that the success of ICL relies on the model's prior knowledge about the query \cite{xie2022incontext,li-etal-2024-language}.
Therefore, we assume that queries that can be enhanced by ICL (\zoney) are more learnable than those unsolvable by ICL (\zonez) but also more valuable for learning than those already solvable by DP (\zonex). 
Motivated by this, we proposed a ZPD-based curriculum learning algorithm for fine-tuning.

\begin{algorithm}[t]
\caption{ZPD-based Curriculum}
\label{alg:DCQGFramework}
\small
\setstretch{1.2}
    \begin{algorithmic}[1]
    \Require Training set $\mathcal{D}$, model $\mathcal{M}$, correct probability with DP $p^{\varnothing}$ and with ICL $p^{c}$, bucket $k$, epoch $e$
    \Ensure  Trained model $\mathcal{M}^{*}$ 
        \State  $\mathcal{D}^{*} \leftarrow $ Sort($\mathcal{D}$, $p^{c}_{i}-p^{\varnothing}_{i} $)
        \State $\{ \mathcal{D}_{1}, ..., \mathcal{D}_{n} \} \leftarrow \mathbf{SplitData}(\mathcal{D}^{*})$ ; $\mathcal{D}_{train} \leftarrow \varnothing$
        \For {$i=1, i\leq k, i$++}
            \State $\mathcal{D}_{train} \leftarrow \mathcal{D}_{train} \cup \mathcal{D}_{i}$ \Comment{Update training set}
            \For {$j=1, j\leq e, j$++}
            \State $\mathbf{Train}(\mathcal{M}, \mathcal{D}_{train})$;
            \EndFor
        \EndFor
    % \State \textbf{return} $\mathcal{M}^{*}$
    \end{algorithmic}
    \label{algo:cl_scheduler}
\end{algorithm}

\noindent \textbf{Approach.} 
Typically, curriculum learning consists of a ranking algorithm, which sorts examples according to a certain criterion, and a scheduling algorithm, which sequences examples for training. 
In our approach, we rank training examples according to $p^{c} - p^{\varnothing}$ (Eq. \ref{eq:sel_icl}), which represents the learning gain brought by ICL. 
For scheduling, we employ the baby-step algorithm \cite{spitkovsky2010baby}, which splits examples into buckets and accumulatively introduces new buckets. The overall process is outlined in Algorithm \ref{algo:cl_scheduler}.

\begin{figure}[t]
    \centering
    \includegraphics[width=\columnwidth]{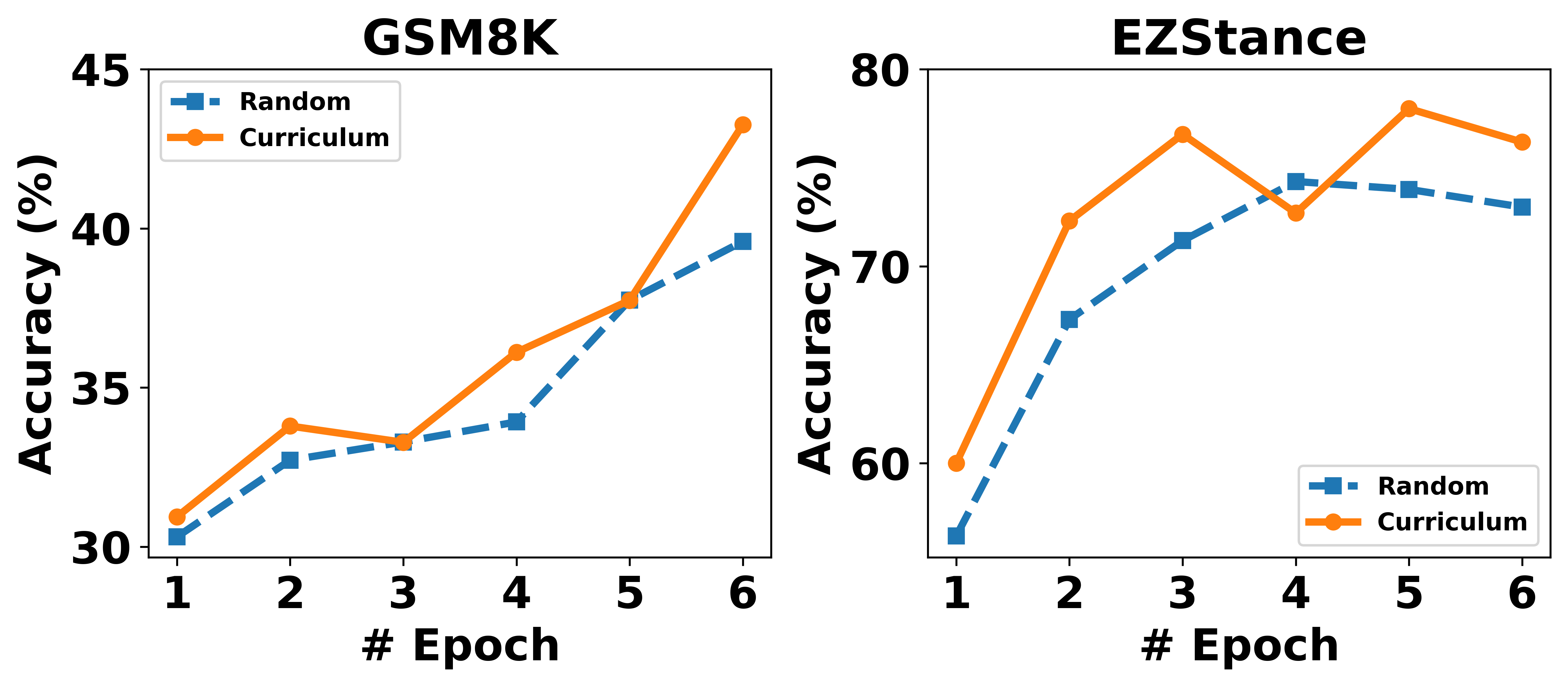}
    \caption{Comparison between random and our \textsc{ZPD}-based curriculum on two datasets.}
    \label{fig:curriculum}
\end{figure}

\noindent \textbf{Results and Analysis.}
We compare our algorithm against a random baseline.
Although simple, random is the most widely used baseline in practice and is not necessarily a weak one, as many curriculum strategies fail to outperform it in language modeling \cite{campos2021curriculum}.
% \TODO{Random is not a weak baseline.}
We fine-tune the \texttt{LLaMA-8B-Instruct} model separately using the two methods with the same scheduler for $6$ epochs. 
See experimental details in Appendix \ref{sec:app_finetune}. 
As shown in Figure \ref{fig:curriculum}, our curriculum results in faster convergence and improved performance in most cases.
To understand why it works, we analyze the training loss of examples in different zones.
Specifically, we compute the \textbf{mean} and \textbf{variance} of each example's loss across epochs. 
The two metrics reflect the convergence behavior of individual examples: a higher mean indicates the example is harder to learn, while a higher variance indicates the model is ambiguous about the example \cite{swayamdipta-etal-2020-dataset}.

\begin{figure}[t]
    \centering
    \includegraphics[width=\columnwidth]{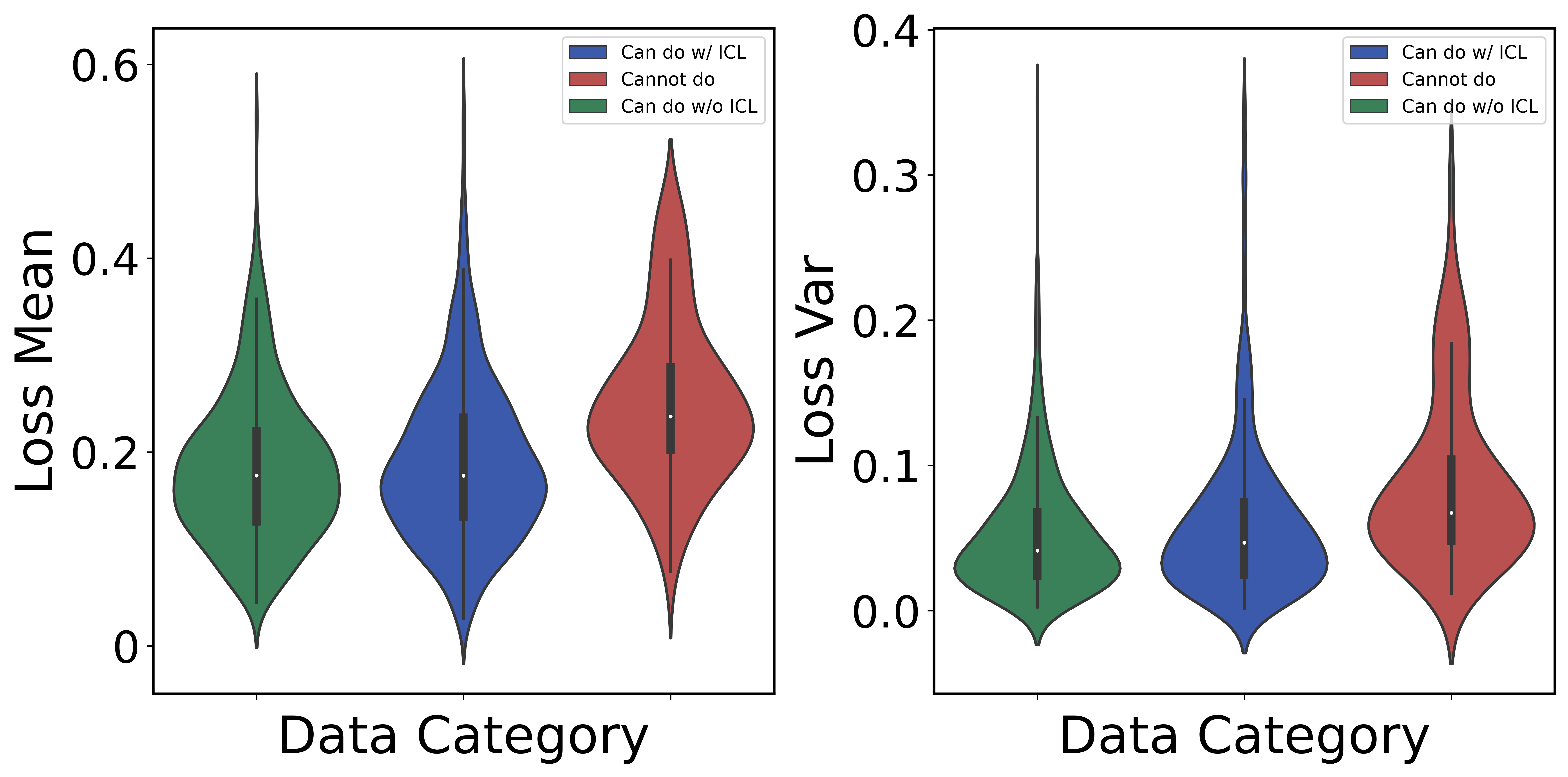}
    \caption{Mean and variance of training loss for queries in different zones. Results are computed over 5 epochs.}
    \label{fig:training_loss}
\end{figure}

For fair analysis, we fine-tune a new \texttt{LLaMA-8B-Instruct} model on the GSM8K dataset for $5$ epochs without any curriculum. 
Figure \ref{fig:training_loss} shows the loss information.
We found \emph{consistent learnability} between in-context learning and fine-tuning scenarios: 
examples in \zonez are the hardest to learn, 
followed by \zoney
\footnote{(Since we use sub-optimal Oracle demonstrations, some \zoney examples are not recalled and misclassified into \zonez. As a result, the actual loss value of \zoney data tends to be slightly closer to \zonez.)}, 
and lastly \zonex. 
This confirms that our curriculum works as expected: prioritizing examples that are learnable and informative (not yet learned).  
Such a strategy has been shown effective for various tasks and model architectures \cite{mindermann2022prioritized,fan2023irreducible}, and our framework provides a new way to discover these examples.

\section{Conclusion}
This work presents a novel framework based on the Zone of Proximal Development (ZPD) theory to analyze the ICL behaviors of LLMs. 
We thoroughly discuss the formalization, measurement, prediction, and application of ZPD in LLMs. 
Our framework serves as an effective tool for understanding the potential, limitations, and complex dynamics of ICL. 
Furthermore, we demonstrate its applicability in both inference and training scenarios.

\section*{Limitations}
We discuss the limitations of this work from the following aspects.
First, due to the unavailability of optimal in-context demonstrations, we can only approximate the ZPD of LLMs, which is a lower bound of the model’s actual in-context learnability. 
This challenge is as nuanced and complex as understanding human learning: one can never precisely measure the potential of human learners.
Second, we investigate the ZPD of LLMs in a simplified scenario where we only consider demonstrations as guidance and use basic templates without instructions to minimize confounding factors. 
In practice, ICL is often combined with other prompting strategies, whose influence may warrant further exploration.
Finally, the ZPD is a dynamic range that evolves with the learner’s knowledge development. 
Our framework is designed to measure and leverage an LLM’s current ZPD, but it is less suited to modeling its developing process (e.g., across different checkpoints during pre-training or fine-tuning). 
In the future, more advanced learning analytics approaches, such as knowledge tracing, could be adopted to enhance our framework.

\bibliographystyle{acl_natbib}

\clearpage
\appendix
\newpage
\section{Experimental Setups}\label{sec:app_setup}
\subsection{Inference Setting} \label{sec:app_inference_setting}
The datasets are used under the MIT License and with their intended use.
For models, we use \texttt{LLaMA} checkpoints from Hugging Face Transformers \cite{wolf2020transformers}.
We run experiments with up to $8 \times$ RTX 4090 24G GPUs. e.
Due to memory constraints, we use Float16 precision for inference, with each run taking around 1\textasciitilde4 hours, depending on the model and data size. 
The prompt template for GSM8K and EZstance are in Appendix Table \ref{tab:app_prompt}.
For ICL, we set the number of demonstrations to $8$ following \cite{li2023unified,rubin2021learning}.

\begin{table}[h]
\small
\centering
\begin{tabular}{@{}p{1.5cm}p{6cm}@{}}
\toprule
\textbf{Dataset}                   & \multicolumn{1}{c}{\textbf{Prompt Template}}                                                                    \\ \midrule
\multirow{2}{*}{\textbf{GSM8K}}    & \texttt{Question:} \colorbox{yellow}{\texttt{\{math\_problem\}}}                                                                            \\
                                   & \texttt{Answer:} \colorbox{yellow}{\texttt{\{step\_by\_step\_answer\}.}}                                                                    \\ \midrule
\multirow{3}{*}{\textbf{EZStance}} & \texttt{Text:} \colorbox{yellow}{\texttt{\{sentence\}}}                                                                                           \\
                                   & \texttt{Question: Which stance-"favor," "against," or "neutral"-does the above text express toward} \colorbox{yellow}{\texttt{\{target\}}}? \\
                                   & \texttt{Answer:} \colorbox{yellow}{\texttt{\{stance\}}}.                                                                                    \\ \bottomrule
\end{tabular}
\caption{Prompt templates for the two datasets. \colorbox{yellow}{highlighted} parts are inputs.} \label{tab:app_prompt}
\end{table}

\subsection{Implementation Details of IRT} \label{sec:app_irt_implementation}
\noindent \textbf{Dataset Construction.}
The dataset for the IRT model is built upon LLM outputs. 
First, we construct Oracle demonstrations using the approach described in \cref{sec: measure}. 
Then, we run LLMs using prompts in Appendix Table \ref{tab:app_prompt} 
 in different settings (DP or ICL). 
The outputs are represented as tuples consisting of $<$\texttt{model\_id}, \texttt{example\_id}, \texttt{input}, \texttt{output}, \texttt{setting}, \texttt{label}$>$.  
This results in total $2$ (Direct prompting or ICL setting) $\times$ $M$ (Number of LLMs) $\times$ $N$ (Number of queries) instances, where $M=8$, $N_{\rm {\tiny GSM8K}}=1319, N_{\rm {\tiny EZStance}}=6703$.  
We further split them into 80\% training set, 10\% validation set, and 10\% test set.

\noindent \textbf{Training Setup.}
We set the dimension of latent traits $\boldsymbol{\theta}, \boldsymbol{\alpha}, \boldsymbol{\theta}^{c},$ $\boldsymbol{\alpha}^{c}$ to 32.
Queries are encoded with SBERT (\texttt{paraphrase-mpnet-base-v2}) with an embedding size of 768.
We train all the models for 10 epochs with a learning rate of $2e-4$ and batch size of 16. 
Traditionally, IRT is optimized by marginalized maximum likelihood estimation \cite{chalmers2012mirt}.
However, this does not scale well to large datasets \cite{lalor2023py}.
We follow \citet{gor2024great} to use Adam \cite{kingma2014adam} to optimize our model.
The best model is selected based on the performance on the validation set.

\subsection{Details of Fine-tuning} \label{sec:app_finetune}
We fine-tune \texttt{LLaMA-3-8B-Instruct} to evaluate our curriculum learning algorithm (\cref{App2}).
Since \texttt{LLaMA} models might already be fine-tuned on the training set of GSM8K \cite{zhang2024careful}, we randomly sample 1,000 instances from the test set for fine-tuning and use the remaining 319 instances for evaluation.
The EZStance dataset is curated after the release of \texttt{LLaMA-3} and, therefore, has no such concern.
We sample 5,000 examples from the training set for fine-tuning and directly evaluate the model on the test set. 
With the scheduler in Algorithm \ref{algo:cl_scheduler}, we split the dataset into $3$ buckets and fine-tune the model on each bucket for $2$ epochs with a learning rate of $1e-5$ and batch size of $4$.

\section{Additional Results}\label{sec:app_add_results}
\subsection{Additional Results of IRT}
The accuracy of IRT models is in Table \ref{tab:app_irt_acc}. 
Note that baseline models are not trained on ICL data and therefore their accuracy is not indicative. 
We report it only for the completeness of the results. 
\begin{table}[h]
\setlength\tabcolsep{4pt}
\small
\centering
\begin{tabular}{@{}lcccccc@{}}
\toprule
\multirow{2}{*}{\textbf{Model}} & \multicolumn{3}{c}{\textbf{GSM8K}} & \multicolumn{3}{c}{\textbf{EZStance}} \\ \cmidrule(l){2-7} 
                                & DP      & ICL      & Overall     & DP       & ICL       & Overall      \\ \midrule
\textsc{Irt}$_{\rm 1Pl}$                       & 69.1      & 39.6     & 56.7        &   76.1         &   45.6        &   63.1           \\
\textsc{Irt}$_{\rm 2PL}$                       & 70.4      & 40.2     & 58.7        &   75.3         &  46.4         &   63.0           \\
MIRT                            & 68.9      & 47.2          & 59.8             & 76.6           & 45.9          & 63.5            \\ 
MIRT$_{\textsc{ICL}}$                            & \textbf{77.4}      & \textbf{78.4}     & \textbf{77.9}        &    \textbf{77.0}       & \textbf{68.5}          &   \textbf{72.8}           \\ \bottomrule
\end{tabular}
\caption{Performance (Accuracy $\%$ ) of various IRT models. The best results are in \textbf{bold}.} \label{tab:app_irt_acc}
\end{table}

\subsection{Additional Results of \textsc{SelIcl}}
The result of \textsc{SelIcl} on the EZStance dataset is in Appendix Figure \ref{fig:ezstance_sel_icl}.

\begin{figure*}[h]
    \centering
    \includegraphics[width=\textwidth]{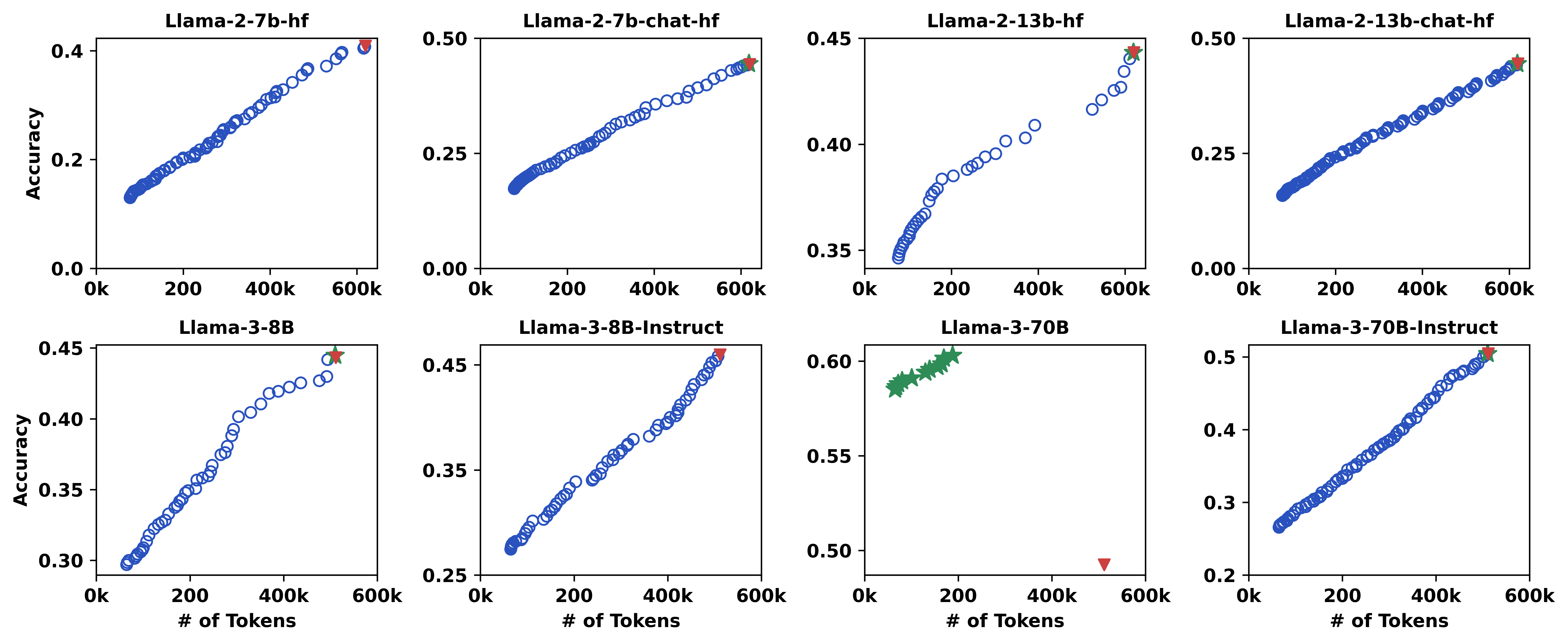}
    \caption{Results of \textsc{SelIcl} on EZStance. See detailed explanations in Figure \ref{fig:gsm8k_sel_icl}.}
    \label{fig:ezstance_sel_icl}
\end{figure*}

\end{document}